\documentclass[11pt,journal]{IEEEtran}

\usepackage{graphicx}
\usepackage{amsmath}
\usepackage{amssymb}
\usepackage{booktabs}
\usepackage{pifont}
\usepackage{xcolor}

\usepackage[pagebackref,breaklinks,colorlinks]{hyperref}

\def\cvprPaperID{7450} 
\def\confName{CVPR}
\def\confYear{2022}

\begin{document}

\title{TGFuse: An Infrared and Visible Image Fusion Approach Based on Transformer and Generative Adversarial Network}

\author{Dongyu Rao,
        Xiao-Jun~Wu,
        Tianyang~Xu
\IEEEcompsocitemizethanks{\IEEEcompsocthanksitem D. Rao and X.-J. Wu (\textsl{Corresponding author}) are with the School of Artificial Intelligence and Computer Science, Jiangnan University, Wuxi, China. (e-mail: raodongyu@163.com, wu\_xiaojun@jiangnan.edu.cn).
\IEEEcompsocthanksitem {T. Xu is with the School of Artificial Intelligence and Computer Science, Jiangnan University, Wuxi 214122, P.R. China and the Centre for Vision, Speech and Signal Processing, University of Surrey, Guildford, GU2 7XH, UK. (e-mail: tianyang\_xu@163.com)}}
}
\maketitle

\begin{abstract}
The end-to-end image fusion framework has achieved promising performance, with dedicated convolutional networks aggregating the multi-modal local appearance.
However, long-range dependencies are directly neglected in existing CNN fusion approaches, impeding balancing the entire image-level perception for complex scenario fusion.
In this paper, therefore, we propose an infrared and visible image fusion algorithm based on a lightweight transformer module and adversarial learning.
Inspired by the global interaction power, we use the transformer technique to learn the effective global fusion relations. 
In particular, shallow features extracted by CNN are interacted in the proposed transformer fusion module to refine the fusion relationship within the spatial scope and across channels simultaneously. 
Besides, adversarial learning is designed in the training process to improve the output discrimination via imposing competitive consistency from the inputs, reflecting the specific characteristics in infrared and visible images. 
The experimental performance demonstrates the effectiveness of the proposed modules, with superior improvement against the state-of-the-art, generalising a novel paradigm via transformer and adversarial learning in the fusion task. 
\end{abstract}

\section{Introduction} \label{sec:intro}
With the development of imaging equipment and analysis approaches, multi-modal visual data is emerging rapidly with many practical applications.
In general, image fusion has played an important role in helping human vision to perceive information association between multi-modal data.
Among them, the fusion of infrared and visible images has important applications in military, security, detection ~\cite{sun2019effective} and visual tracking ~\cite{luo2016novel,luo2017image,li2020mdlatlrr, xu2019learning,xu2021adaptive,xu2020accelerated} etc., becoming an important part of image fusion tasks.

\begin{figure}[t]
  \centering
  \includegraphics[width=1\linewidth]{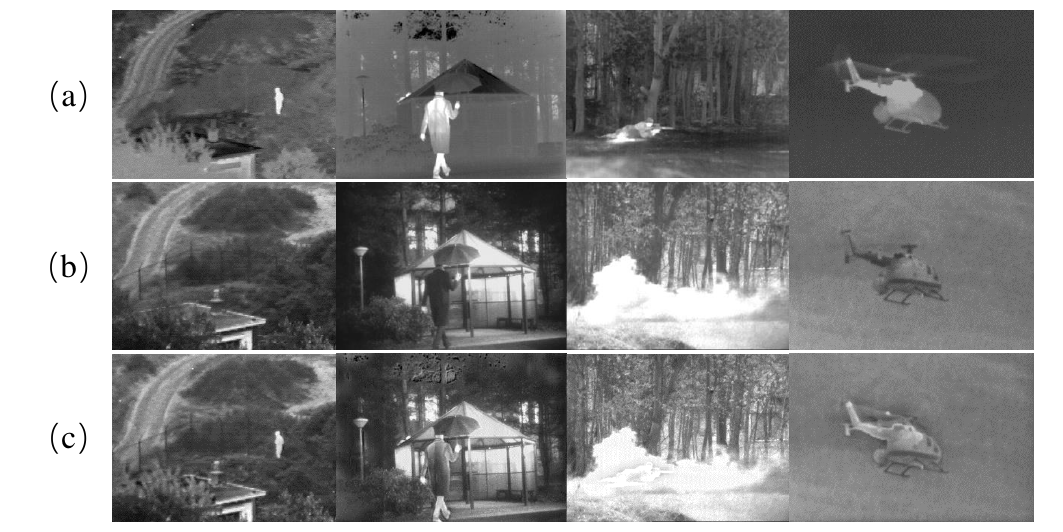}
  \caption{Infrared image (a), visible image (b) and fused image generated by the proposed method (c).}
  \label{fig:ir}
\vspace{0.14in}
\end{figure}
\begin{figure*}
  \centering
  \includegraphics[width=0.8\linewidth]{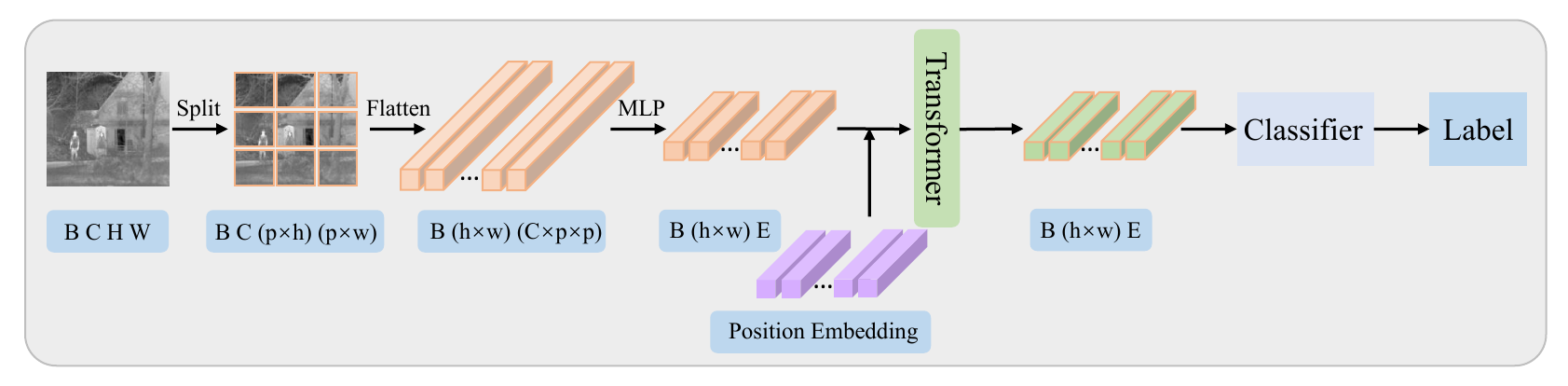}
  \caption{The framework of ViT (Vision Transformer). "B C H W" respectively represent the batch size, channels, height and width. "p" means patch size. "h w" is the number of patches in height and width. "E" is the reduced dimension.}
  \label{fig:vit}
\end{figure*}
\setlength{\textfloatsep}{2ex}
In order to design a natural and efficient image fusion algorithm, researchers have developed many fusion algorithms on the basis of traditional image processing.
Firstly, the fusion algorithms based on multi-scale transformation are proposed ~\cite{mertens2007exposure, zhang1999categorization,chen2018new,li2011no}, which applied traditional image processing methods to image fusion. 
Subsequently, fusion algorithms based on sparse / low-rank representation were applied ~\cite{chen2014image, nejati2015multi,zheng2006nearest}.
These algorithms use specific image processing methods to obtain image representations, and obtain the output images by fusing the image representations.
However, the image features obtained by these methods are relatively less salient.
Most of the fusion methods also require complex designs, so that the fusion results usually introduce a large amount of noise.
With the development of deep learning, image fusion methods based on convolutional neural networks have become the mainstream of the topic ~\cite{liu2017multi,li2021rfn}. 
However, since most image fusion tasks are unsupervised, the supervised end-to-end training framework is not suitable for training fusion tasks.
Drawing on this, some fusion algorithms ~\cite{li2019infrared} used large-scale pre-trained networks to extract image features. 
However, the pre-trained network is mostly used for classification tasks, and the extracted features cannot meet the requirements of the fusion task.
Subsequently, Li et al. ~\cite{li2018densefuse,li2020nestfuse} proposed a fusion algorithm based on an encoder-decoder network, using ordinary data sets for encoder-decoder training. This method makes the fusion task get rid of the dependence on multi-modal data sets. But this also makes it unable to effectively learn specific tasks.
In order to obtain better performance for specific fusion tasks, the end-to-end image fusion methods ~\cite{ma2019fusiongan,zhang2020ifcnn,fu2021image} are proposed to learn more targeted network parameters through a specific network structure and loss function.
This method is dedicated to training fusion tasks, which can usually achieve better fusion results.
However, this puts forward higher requirements for the representative ability of the network and the effectiveness of the fusion method.
At present, the end-to-end fusion algorithm mainly uses a convolutional neural network for feature extraction and achieves the fusion effect.
However, due to the characteristics of CNN, this process usually ignores the global dependency infusion.

In order to solve the problem of global dependence and effective integration, we propose an infrared and visible image fusion algorithm based on the lightweight transformer and adversarial learning.
Our method uses a general visual transformer for image spatial relationship learning. 
In particular, we propose a novel cross-channel transformer model to learn the channel relationship.
The composite transformer fusion module has learned the global fusion relationship with space and channels. 
In addition, adversarial learning is introduced in the training process. 
We use two discriminators (infrared and fused image, visible and fused image) for adversarial training respectively. 
This allows the fused image to obtain higher-quality infrared and visible image characteristics.


The proposed method mainly has the following three innovations:
\begin{itemize}
\item A channel-token transformer is proposed to explore the channel relationships, which is effectively applied in the fusion method.

\item A transformer module is designed to achieve global fusion relationship learning in complex scenarios.

\item Adversarial learning is introduced into the training process. The discriminator of the two modalities introduces the characteristics of different modalities to the fused image to improve the fusion effect.
\end{itemize}

\section{Related work} \label{sec:re_work}
Although traditional methods are well investigated \cite{wang2003initial, sun2011quantum}, deep learning based methods are mainly discussed in this paper.
\subsection{Image Fusion Method Based on Deep Learning}
The fusion algorithm based on deep learning has shown excellent performance in infrared and visible image fusion, multi-focus image fusion and medical image fusion, etc. 
Li et al. ~\cite{li2018infrared,li2019infrared} used a pre-trained neural network to extract image features and used them for image fusion weight calculation.
This is a preliminary combination of neural network and image fusion tasks.
In order to obtain the depth features suitable for reconstructing images, Li et al. ~\cite{li2018densefuse} first proposed an algorithm based on an auto-encoder network. In the absence of specific data, the algorithm can also achieve a good fusion effect.
With the advancement of visual data collection equipment, some large-scale multi-mode data sets have appeared, so end-to-end fusion algorithms ~\cite{xu2020u2fusion,ma2020infrared} have received more attention and applications. 
This end-to-end fusion algorithm based on convolutional neural networks achieves better performance on a single task.
But it still has some limitations, such as the spatial limitation of the fusion method based on a convolutional neural network.
In this paper, the proposed method is an end-to-end image fusion algorithm.
But compared to the CNN-based fusion network, we expand the network structure of the end-to-end algorithm and introduce the transformer that focuses on building global relationships into the fusion module. Our algorithm opens up new ideas in the design of fusion methods.

\subsection{Generative Adversarial Network}
\begin{figure*}[t]
  \centering
  \includegraphics[width=0.8\linewidth]{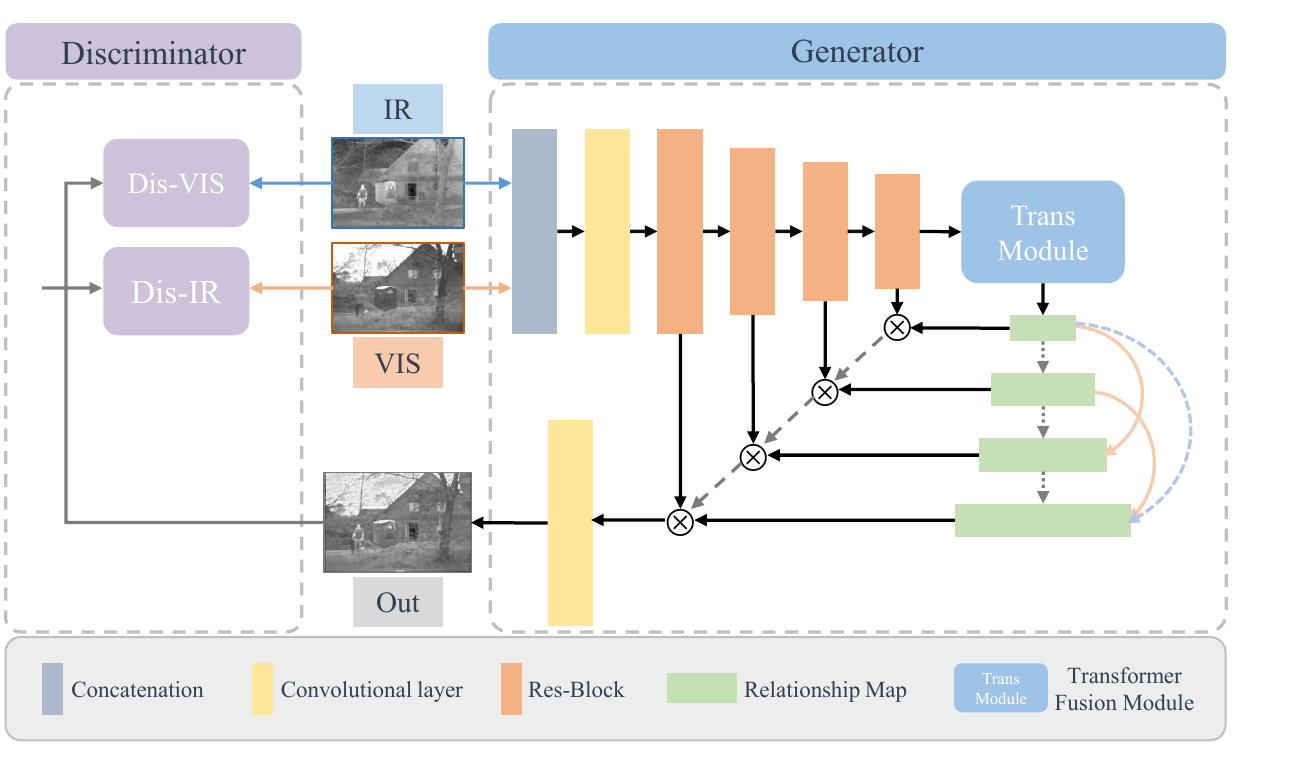}
  \caption{The framework of our method.}
  \label{fig:framework}
\end{figure*}
\begin{figure}[t]
  \centering
  \includegraphics[width=0.8\linewidth]{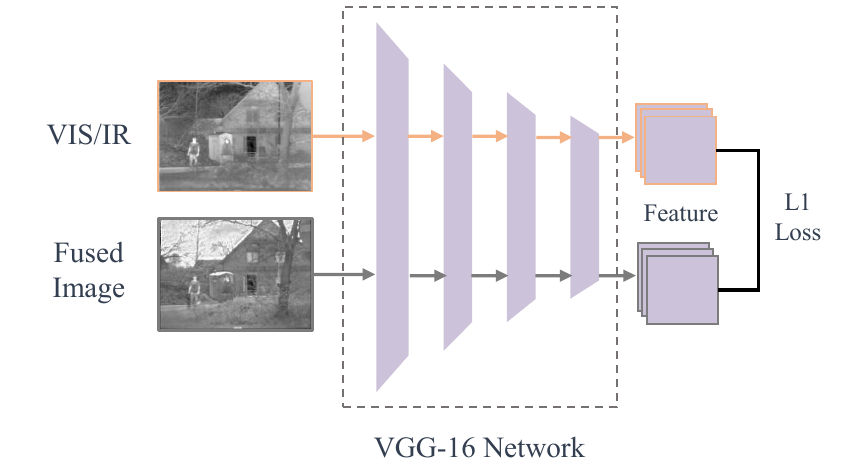}
  \caption{The framework of discriminator.}
  \label{fig:dis}
\end{figure}

A generative adversarial network (GAN) is an algorithm that obtains high-quality generated images by training two networks against each other.
Goodfellow et al. ~\cite{goodfellow2014generative} first proposed the idea of a generative adversarial network. 
The generator generates an image, and the discriminator determines whether the input image is a real image (True) or a generated image (False).
Subsequently, many improvements based on the original GAN focused on speeding up the training of the network and improving the quality of the generated images ~\cite{mao2017least,zhao2017energy,berthelot2017began}.
These improvements also help GAN gain a wider range of applications ~\cite{liang2021high, liu2021pd, xia2021tedigan}.
Methods based on GAN are also widely used in image generation tasks ~\cite{karras2019style,zhu2017unpaired}.
There are already some image fusion methods based on GAN~\cite{ma2019fusiongan, fu2021image}. 
Adversarial learning is an important part of our approach. It improves the infrared and visible image characteristics in the fusion result by obtaining competitive consistency from the inputs.
However, we abandon the discriminator of the classification mode and use the difference in the feature level to promote the fused image to have more infrared or visible image information.

\subsection{Visual Transformer}
\begin{figure}[t]
  \centering
  \includegraphics[width=0.8\linewidth]{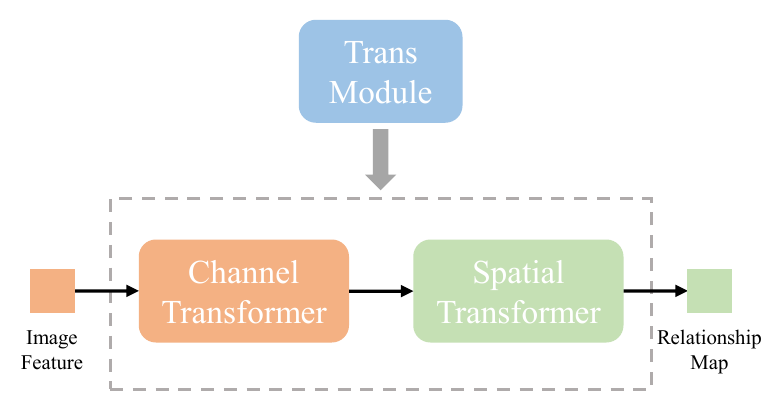}
  \caption{The framework of transformer fusion module.}
  \label{fig:trans-module}
\end{figure}
\setlength{\textfloatsep}{2ex}
The transformer is a model based on a pure attention mechanism~\cite{vaswani2017attention}. Its success in natural language processing inspires its application in computer vision. 
Due to the long-range dependence of the transformer in processing input, the visual transformer also has the ability to pay attention to the global relationship in image tasks.
As a pioneering work of visual transformer, Dosovitskiy et al.~\cite{dosovitskiy2020image} 
proposed ViT (Vision Transformer) for image classification tasks (Figure.\ref{fig:vit}). This is a simple and effective application of transformer in visual tasks.
Subsequently, Chen et al.~\cite{chen2021pre} proposed a multi-task model based on the transformer, which achieved good results on multiple low-level visual tasks. The global spatial dependence of transformers has gained many applications in the field of computer vision. 
Inspired by the characteristics of the transformer, we pay attention to the global correlation of images space and channels during the fusion process.
We propose a new transformer model that focuses on channel relationships and applies it in the field of image fusion. 
Compared with the general transformer, our transformer fusion module is a lightweight model. This is a new exploration of transformer applications.
\vspace{-0.1in}
\section{Proposed Method}
\subsection{The Framework of Network}
\begin{figure*}
  \centering
  \includegraphics[width=0.9\linewidth]{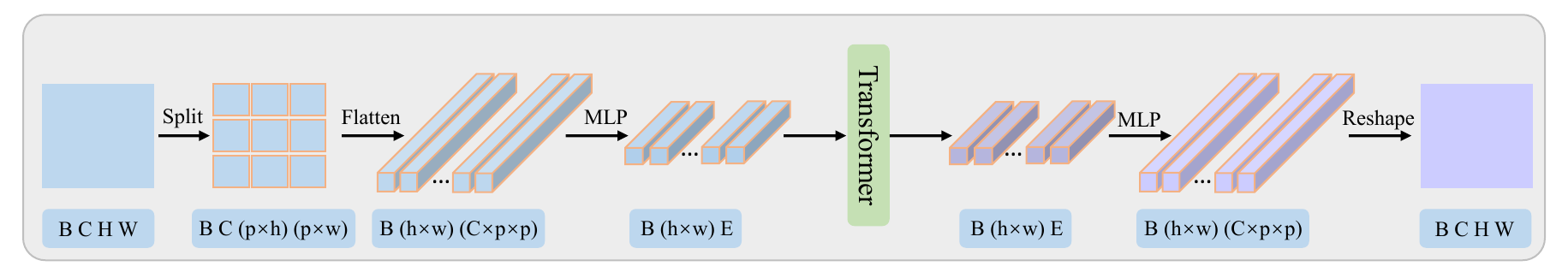}
  \caption{The framework of spatial transformer.}
  \label{fig:spatial}
\end{figure*}
\begin{figure*}
  \centering
  \includegraphics[width=0.9\linewidth]{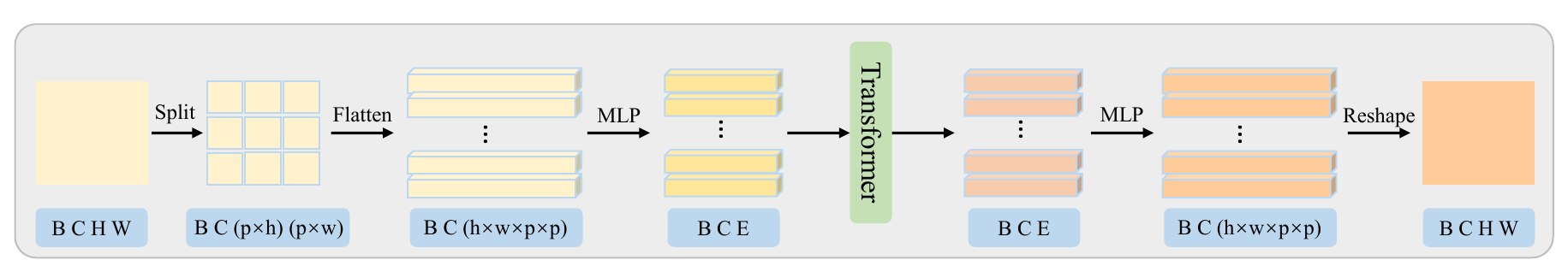}
  \caption{The framework of channel transformer.}
  \label{fig:channel}
\end{figure*}

As shown in Figure. \ref{fig:framework}, our model is mainly composed of two parts: one transformer-based generator and two discriminators. 
Typically, the fused image is obtained by the generator. 
Then, the output is refined during the adversarial learning between the generator and the discriminator.

\textbf{Generator.}
The generator is used for the generation of the fused image.
After the source images are merged in the channel dimension, the initial feature extraction is performed through the convolutional neural network. 
The mixed CNN features are input to the transformer fusion module to learn global fusion relations.
Taking into account the consumption of computing resources and representation of features, three downsampling operators are added before the transformer fusion module. 
The fusion relationship learned in this process is up-sampled to different scales and multiplied by the corresponding features to achieve the preliminary result.
The fusion features of different scales are up-sampled to the original image size and then superimposed to obtain the final fusion result.

\textbf{Discriminator.}
The discriminator is used to refine the perception quality of the fused image.
We set up two discriminators: fused image and infrared image ("Dis-IR"), fused image and visible image ("Dis-VIS"). 
These two discriminators provide high-resolution details of the visible image and a significant part of the infrared image for the fused image.
The pre-trained VGG-16 network is used as the discriminator, which can be further fine-tuned during training.
The network is shown in Figure.\ref{fig:dis}.
Taking the visible image discriminator ("Dis-VIS") as an example, the fused image and the visible image are separately input into the VGG-16 network to extract features. 
We calculate the L1 loss between the two features so that the fused image approximates the visible image from the context perspective.
According to the number of downsampling, VGG-16 is divided into 4 layers. Different layers have different feature depths and different feature shapes.
Inspired by Johnson et al.~\cite{johnson2016perceptual}, we use the features of different depths extracted by VGG-16 to distinguish between infrared and visible features.
The infrared discriminator uses the features of the fourth layer of VGG-16 to retain more saliency information.
While the visible discriminator uses the features of the first layer of VGG-16 to retain more detailed information.

In the training stage, source images are input to the generator to obtain the preliminary fused image.
The preliminary fused image then passes through two discriminators with the effect of the fused image being fed back through the loss function.
The above two steps are performed alternately to realize the confrontation training between the generator and the discriminator.
Finally, we get a generator with an ideal generation effect to achieve the purpose of image fusion. 

\subsection{The Transformer Fusion Module}
As shown in Figure. \ref{fig:trans-module}, the transformer fusion module consists of two parts: general transformer ("spatial transformer") and cross-channel transformer ("channel transformer"). This helps us to obtain a more comprehensive global integration relationship. 

\begin{figure*}
  \centering
  \includegraphics[width=0.8\linewidth]{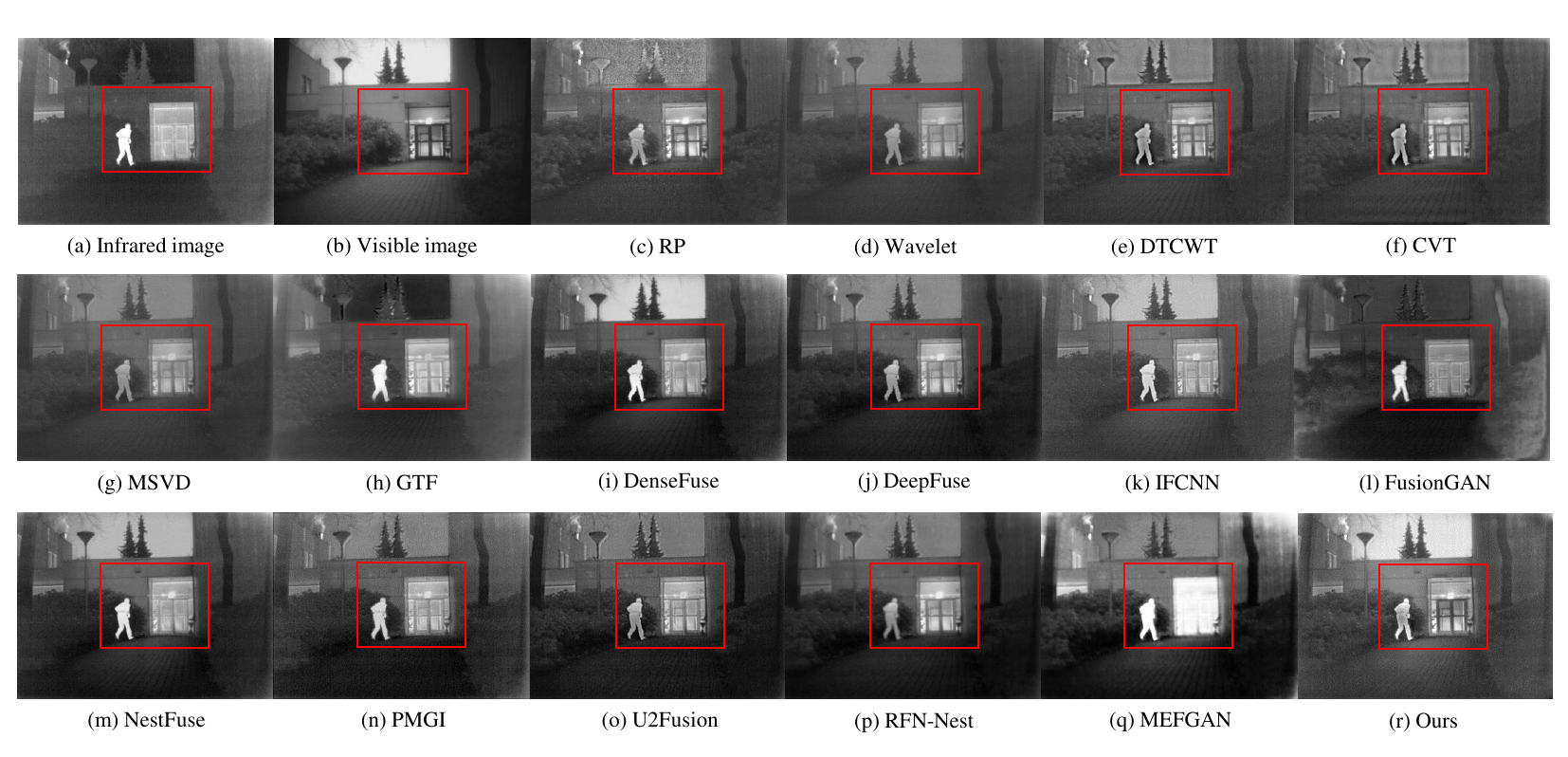}
  \caption{Infrared and visible image fusion experiment on “human” images.}
  \label{fig:result}
\end{figure*}

\textbf{Spatial Transformer}
As shown in Figure. \ref{fig:vit}, the image is divided into blocks and stretched into vectors, where "p" means patch size, "w" and "h" respectively represent the number of image blocks in the width and height dimensions of the image, "E" is the reduced dimension.
Then, the vector group enters the transformer model for relation learning. 
The number of image blocks is used to learn the global relationship of the image. 
Therefore, we consider that the general transformer mainly learns the global spatial relationship between image patches.
Inspired by the transformer-based low-level image task, we build a spatial transformer for the fusion task. 
As shown in Figure. \ref{fig:spatial}, the spatial transformer is basically the same as the first half of ViT (Figure. \ref{fig:vit}). 
The difference is that we cancelled the addition of position embedding, and subsequent experiments also proved the rationality and effectiveness of this operation. 
In addition, when restoring from the vector group to the image, we compress the channel dimension, so that we get a relationship map with a channel number of 1.
This corresponds to the spatial relationship of the image we obtained, avoiding the interference of other dimensional relationships.

\textbf{Channel Transformer}
For image fusion tasks, we believe that the cross-channel relationship of images also plays an important role in fusion.
Therefore, we propose a new cross-channel transformer model, which learns the correlation of information across the channel dimension.
In the new transformer module, the number of tokens input to the encoder has changed from the number of image blocks to the number of image channels.
Since position embedding is not required to provide category information in the image generation task, we have removed position embedding, which also makes the size of the input image more flexible. 
The channel transformer is also a structure similar to the spatial transformer.
The main difference is that we change the object modelled by the transformer from the spatial relationship of the image block to the channel relationship. 
In this specific implementation, we use the number of channels as the token number, which is a simple but effective operation.
Through two kinds of the transformer, we can get the relation mapping for the image fusion task.

\textbf{Composite Transformer}
The transformer of the two modes is combined into a transformer fusion module, which enables our fusion model to simultaneously learn spatial and channel relationships with global correlation.
Through experiments, we find that using a channel transformer first and then using a spatial transformer can achieve better results.
This shows that the combination of these two fusion modules is used to learn the coefficients that are more suitable for the fusion of infrared and visible images.

\subsection{Loss Function}
Previous image fusion algorithms based on deep learning usually use multiple loss functions to optimize the fused image from different perspectives during training. 
But this causes mutual conflict among loss functions.
Inspired by ~\cite{hou2020vif}, we make improvements on the basis of the SSIM loss.
A single loss function achieves a good fusion effect and avoids the problem of entanglement of multiple loss functions.

SSIM~\cite{wang2004image} is a measure of structural similarity between images. 
As shown in Eq. \eqref{eq:ssim}, X, Y represent two images respectively. 
$\mu$ and $\sigma$ stand for mean and standard deviation respectively. 
$\sigma_{XY}$ means the covariance between X and Y. 
$C_1$ and $C_2$ are stability coefficients.
\begin{equation}\label{eq:ssim}\small
SSIM(X, Y) = \frac{(2\mu_X\mu_Y + C_1)(2\sigma_{XY} + C_2)}{({\mu_X}^2 + {\mu_Y}^2 + C_1)({\sigma_X}^2 + {\sigma_Y}^2 + C_2)}
\end{equation}

Variance reflects the contrast of the image, and an image with high contrast is more helpful for the human visual system to capture information. 
As shown in Eq. \eqref{eq:var}, \textit{M} and \textit{N} are the image size in the horizontal and vertical directions respectively. 
$\mu$ represents the mean of the image. 
We use variance as the standard and choose one as the reference image from infrared and visible images.
The structural similarity between the fused image and the reference image is calculated, so that the fused image gradually approaches the reference image during the optimization process. 
This operation allows the fusion result to better obtain the important information from the infrared or visible image.
\begin{equation}\label{eq:var}
\sigma^2(X) = \frac{\mathop\Sigma\limits_{i=0}\limits^{M-1}\mathop\Sigma\limits_{j=0}\limits^{N-1}[X(i, j) - \mu]^2}{MN}
\end{equation}

In Eq. \eqref{eq:varssim}, $Var_-SSIM$ calculates the structural similarity of the divided image. 
$\sigma^2$ is the variance of the image. 
$I_X$ and $I_Y$ represent two source images respectively. 
$I_F$ means a fused image. 
$W$ is the number of image blocks after division, and the size of each image block is set to 11$\times$11.
Image segmentation is achieved through sliding windows.
Through the sliding window, the fused image can well coordinate the consistency between different image blocks.
The calculation of the loss function is shown in Eq. \eqref{eq:loss}.

\begin{equation}\label{eq:varssim}\footnotesize
Var_-SSIM(I_X, I_Y, I_F|W) = 
\left\{
             \begin{array}{lr}
             SSIM(I_X, I_F), \\
             if \sigma^2(X) > \sigma^2(Y) &  \\
             SSIM(I_Y, I_F), \\
             if \sigma^2(Y) >= \sigma^2(X) &  
             \end{array}
\right.
\end{equation}

\begin{equation}\label{eq:loss}\small
L_{var_-SSIM} = 1 - \frac{1}{N}\mathop\Sigma\limits^N\limits_{W=1} Var_-SSIM(I_X, I_Y, I_F|W)
\end{equation}

\section{Experiments}
\subsection{Setup}
\textbf{Datasets.}
In the training phase, 40,000 pairs of corresponding infrared and visible images are selected as the training data from the KAIST ~\cite{hwang2015multispectral} data set. 
KAIST data set is a pedestrian data set containing various general scenes of campus, street and countryside.
Each picture contains a visible image and a corresponding infrared image. 
At present, some end-to-end image fusion algorithms~\cite{li2021rfn} use it as training data.
The training image size is set to 256×256 pixels.
In the testing phase, we use 10 pairs of images from the test image of ~\cite{li2018densefuse} as the test set. 
The size of the test data is arbitrary (generally not more than 2048×2048 pixels).

\textbf{Hyper-Parameters.}
In the training phase, we choose Adam as the optimizer and the learning rate is set to a constant of 0.0001. 
Training data includes 40,000 pairs of images and batch size is set to 16.
Complete training requires 20 epochs. 
Inspired by ~\cite{dosovitskiy2020image, chen2021pre}, we chose fixed values for some parameters in the transformer fusion module.
The patch size of the spatial transformer and channel transformer is set to 4 and 16 respectively. 
Taking into account the different dimensions of the data processed by a spatial transformer and channel transformer, the embedding dimensions are set to 2048 and 128 respectively.
Our model is implemented with NVIDIA TITAN Xp and Pytorch.

\begin{table*}[!t]
\small
\caption{Quantitative evaluation results of infrared and visible image fusion tasks. The best three results are highlighted in {\color{red}{\textbf{red}}}, {\color{brown}{\textbf{brown}}} and {\color{blue}{\textbf{blue}}} fonts.}
\label{table:result}
\centering
\begin{tabular}{cccccccccc}
\hline
    Method          & SF               & EN              & Q$_{abf}$            & FMI$_w$          & MS-SSIM        & FMI$_{pixel}$      & MI               & SD               & VIF             \\
\hline
RP~\cite{toet1989image}            & \color{red}{\textbf{12.7249}} & 6.5397          & 0.4341          & 0.3831          & 0.8404          & 0.8929          & 13.0794          & 63.2427          & 0.6420          \\
Wavelet~\cite{chipman1995wavelets}       & 6.2567           & 6.2454          & 0.3214          & 0.4183          & 0.8598          & 0.9096          & 12.4907          & 52.2292          & 0.2921          \\
DTCWT~\cite{lewis2007pixel}         & 11.1296          & 6.4791          & \color{brown}{\textbf{0.5258}} & 0.4419          & 0.9053          & \color{brown}{\textbf{0.9186}} & 12.9583          & 60.1138          & 0.5986          \\
CVT~\cite{nencini2007remote}           & 11.1129          & 6.4989          & 0.4936          & 0.4240          & 0.8963          & \color{blue}{\textbf{0.9156}} & 12.9979          & 60.4005          & 0.5930          \\
MSVD~\cite{naidu2011image}          & 8.5538           & 6.2807          & 0.3328          & 0.2828          & 0.8652          & 0.9036          & 12.5613          & 52.9853          & 0.3031          \\
GTF~\cite{ma2016infrared}           & 9.5022           & 6.5781          & 0.4400          & \color{red}{\textbf{0.4494}} & 0.8169          & 0.9056          & 13.1562          & 66.0773          & 0.4071          \\
DenseFuse~\cite{li2018densefuse}     & 9.3238           & 6.8526          & 0.4735          & 0.4389          & 0.8692          & 0.9061          & 13.7053          & \color{blue}{\textbf{81.7283}} & 0.6875          \\
DeepFuse~\cite{prabhakar2017deepfuse}      & 8.3500           & 6.6102          & 0.3847          & 0.4214          & \color{blue}{\textbf{0.9138}} & 0.9041          & 13.2205          & 66.8872          & 0.5752          \\
IFCNN~\cite{zhang2020ifcnn}         & \color{brown}{\textbf{11.8590}} & 6.6454          & 0.4962          & 0.4052          & 0.9129          & 0.9007          & 13.2909          & 73.7053          & 0.6090          \\
FusionGAN~\cite{ma2019fusiongan}     & 8.0476           & 6.5409          & 0.2682          & 0.4083          & 0.6135          & 0.8875          & 13.0817          & 61.6339          & 0.4928          \\
NestFuse~\cite{li2020nestfuse}  & 9.7807           & \color{blue}{\textbf{6.8745}} & \color{blue}{\textbf{0.5011}} & \color{brown}{\textbf{0.4483}} & 0.8817          & 0.9025          & \color{blue}{\textbf{13.7491}} & \color{brown}{\textbf{83.0530}} & 0.7195          \\
PMGI~\cite{zhang2020rethinking}          & 8.7195           & 6.8688          & 0.3787          & 0.4018          & 0.8684          & 0.9001          & 13.7376          & 69.2364          & 0.6904          \\
U2Fusion~\cite{xu2020u2fusion}      & 11.0368          & 6.7227          & 0.3934          & 0.3594          & \color{brown}{\textbf{0.9147}} & 0.8942          & 13.4453          & 66.5035          & \color{brown}{\textbf{0.7680}} \\
RFN-Nest~\cite{li2021rfn}      & 5.8457           & 6.7274          & 0.3292          & 0.3052          & 0.8959          & 0.9063          & 13.4547          & 67.8765          & 0.5404          \\
MEFGAN~\cite{xu2020mef}        & 7.8481           & \color{brown}{\textbf{6.9727}} & 0.2076          & 0.1826          & 0.6709          & 0.8844          & \color{brown}{\textbf{13.9454}} & 43.7332          & \color{blue}{\textbf{0.7330}} \\
TGFuse(ours)          & \color{blue}{\textbf{11.3149}} & \color{red}{\textbf{6.9838}} & \color{red}{\textbf{0.5863}} & \color{blue}{\textbf{0.4452}} & \color{red}{\textbf{0.9160}} & \color{red}{\textbf{0.9219}} & \color{red}{\textbf{13.9676}} & \color{red}{\textbf{94.7203}} & \color{red}{\textbf{0.7746}}
  \\
\hline
\end{tabular}
\end{table*}

\begin{table*}[!t]
\caption{The objective evaluation on whether to use GAN. The best results are highlighted in \textbf{bold} fonts.}\small
\label{table:gan}
\centering
\begin{tabular}{cccccccccc}
\hline
& SF               & EN              & Q$_{abf}$            & FMI$_w$          & MS-SSIM        & FMI$_{pixel}$      & MI               & SD               & VIF             \\
\hline
w/o GAN & 11.2253 & 6.9547 & 0.5794 & 0.4425 & \textbf{0.9240}   & 0.9212     & 13.9094 & 92.4749 & \textbf{0.7870} \\
GAN     & \textbf{11.3149} & \textbf{6.9838} & \textbf{0.5863} & \textbf{0.4452} & 0.9160   & \textbf{0.9219}     & \textbf{13.9676} & \textbf{94.7203} & 0.7746 \\
\hline
\end{tabular}
\end{table*}

\begin{table*}[!t]
\caption{The objective evaluation on different transformer fusion method. The best results are highlighted in \textbf{bold} fonts.}
\label{table:trans}\small
\centering
\begin{tabular}{cccccccccc}
\hline
& SF               & EN              & Q$_{abf}$            & FMI$_w$          & MS-SSIM        & FMI$_{pixel}$      & MI               & SD               & VIF             \\
\hline
Spatial         & 10.8364          & 6.8665          & 0.5491          & 0.4281          & \textbf{0.9337} & 0.9173          & 13.7330          & 86.2626          & 0.7247          \\
Channel         & 11.1283          & 6.9520          & 0.5622          & 0.4328          & 0.9107          & 0.9169          & 13.9040          & 91.2356          & 0.7417          \\
Spatial+Channel & 10.8808          & 6.9161          & 0.5304          & 0.4139          & 0.9172          & 0.9089          & 13.8323          & \textbf{94.6343} & 0.7565          \\
Channel+Spatial & \textbf{11.2253} & \textbf{6.9547} & \textbf{0.5794} & \textbf{0.4425} & 0.9240          & \textbf{0.9212} & \textbf{13.9094} & 92.4749          & \textbf{0.7870}\\
\hline
\end{tabular}
\end{table*}

\begin{table*}[!t]
\caption{The objective evaluation on whether to use position embedding. The best results are highlighted in \textbf{bold} fonts.}
\label{table:pe}\small
\centering
\begin{tabular}{cccccccccc}
\hline
& SF               & EN              & Q$_{abf}$            & FMI$_w$          & MS-SSIM        & FMI$_{pixel}$      & MI               & SD               & VIF             \\
\hline
w/o PE & \textbf{11.2253} & \textbf{6.9547} & \textbf{0.5794} & \textbf{0.4425} & 0.9240          & \textbf{0.9212} & \textbf{13.9094} & 92.4749          & \textbf{0.7870} \\
PE     & 10.8748          & 6.9332          & 0.5522          & 0.4186          & \textbf{0.9340} & 0.9174          & 13.8664          & \textbf{90.5422} & 0.7654 \\
\hline
\end{tabular}
\end{table*}

\begin{table*}[!t]
\caption{The objective evaluation on different encoder layers of transformer. The best results are highlighted in \textbf{bold} fonts.(“\textbf{/}” means training failure)}
\label{table:trans-layer}\small
\centering
\begin{tabular}{cccccccccc}
\hline
& SF               & EN              & Q$_{abf}$            & FMI$_w$          & MS-SSIM        & FMI$_{pixel}$      & MI               & SD               & VIF             \\
\hline
3-layers & \multicolumn{9}{c}{\textbf{/}}                                                                                                                                     \\
4-layers & \textbf{11.2253} & \textbf{6.9547} & \textbf{0.5794} & \textbf{0.4425} & 0.9240          & \textbf{0.9212} & \textbf{13.9094} & \textbf{92.4749} & \textbf{0.7870} \\
5-layers & 11.1740          & 6.8722          & 0.5623          & 0.4209          & \textbf{0.9404} & 0.9198          & 13.7443          & 86.7715          & 0.7539  \\
\hline
\end{tabular}
\end{table*}

\begin{table*}[!t]
\caption{The objective evaluation on different layers of CNN. The best results are highlighted in \textbf{bold} fonts.(“\textbf{/}” means training failure)}
\label{table:cnn-layer}\small
\centering
\begin{tabular}{cccccccccc}
\hline
& SF               & EN              & Q$_{abf}$            & FMI$_w$          & MS-SSIM        & FMI$_{pixel}$      & MI               & SD               & VIF             \\
\hline
2-layers & 10.3438          & 6.7281          & 0.5560          & 0.4314          & 0.9006          & 0.9097          & 13.4562          & \textbf{94.2280} & 0.6862          \\
3-layers & 11.0769          & 6.8959          & 0.5497          & 0.4272          & \textbf{0.9298} & 0.9157          & 13.7919          & 92.5518          & 0.7517          \\
4-layers & \textbf{11.2253} & \textbf{6.9547} & \textbf{0.5794} & \textbf{0.4425} & 0.9240          & \textbf{0.9212} & \textbf{13.9094} & 92.4749          & \textbf{0.7870} \\
5-layers & \multicolumn{9}{c}{\textbf{/}}  \\
\hline
\end{tabular}
\end{table*}

\begin{table*}[!t]
\caption{The objective evaluation on different channels. The best results are highlighted in \textbf{bold} fonts.}
\label{table:cnn-channel}\small
\centering
\begin{tabular}{cccccccccc}
\hline
& SF               & EN              & Q$_{abf}$            & FMI$_w$          & MS-SSIM        & FMI$_{pixel}$      & MI               & SD               & VIF             \\
\hline
32-channels  & 10.6360          & 6.9228          & 0.5715          & 0.4370          & 0.9276          & 0.9206          & 13.8456          & 90.1796          & 0.7061          \\
64-channels  & \textbf{11.2253} & \textbf{6.9547} & \textbf{0.5794} & \textbf{0.4425} & 0.9240          & \textbf{0.9212} & \textbf{13.9094} & \textbf{92.4749} & \textbf{0.7870} \\
128-channels & 11.1181          & 6.9388          & 0.5545          & 0.4142          & \textbf{0.9368} & 0.9163          & 13.8776          & 88.5524          & 0.8069  \\
\hline
\end{tabular}
\end{table*}

\textbf{Compared Methods.}
The proposed method is compared with 15 methods in subjective and objective evaluation, including classic and latest methods. 
These are: Ratio of Low-pass Pyramid (RP) ~\cite{toet1989image}, Wavelet ~\cite{chipman1995wavelets}, Dual-Tree Complex Wavelet Transform (DTCWT) ~\cite{lewis2007pixel}, Curvelet Transform (CVT) ~\cite{nencini2007remote}, Multi-resolution Singular Value Decomposition (MSVD) ~\cite{naidu2011image}, gradient transfer and total variation minimization (GTF) ~\cite{ma2016infrared}, DenseFuse ~\cite{li2018densefuse}, DeepFuse ~\cite{prabhakar2017deepfuse}, a general end-to-end fusion network(IFCNN) ~\cite{zhang2020ifcnn}, FusionGAN ~\cite{ma2019fusiongan}, NestFuse ~\cite{li2020nestfuse}, PMGI~\cite{zhang2020rethinking}, U2Fusion ~\cite{xu2020u2fusion}, RFN-Nest~\cite{li2021rfn}, and MEFGAN~\cite{xu2020mef}, respectively. 

\subsection{Results Analysis}
We use subjective evaluation and objective evaluation to measure the performance of the fusion algorithm.
Subjective evaluation judges whether the fusion result conforms to human visual perception, such as clarity, salient information, etc.
Therefore, the subjective evaluation method puts the fused images obtained by different algorithms together for intuitive visual comparison.

In Figure. \ref{fig:result}, the fusion results of all methods are put together for subjective judgment. Although some methods can achieve a certain fusion effect, it introduces more artificial noise, which affects the acquisition of visual information, such as (c), (d), (e), (f), (g). In contrast, the fusion result produced by the deep learning method is more in line with human vision. 
Most methods based on deep learning can maintain the basic environmental information of the visible image and the salient human of the infrared image at the same time.
Compared with other methods, our method not only highlights the infrared information of the person in the red frame but also maintains the visible details of the door. The sky as the background also retains the high-resolution visible scene. 
Such a fused image is friendly and easy to accept information for human vision.

There are many different evaluation indicators for objective evaluation. We have selected nine common evaluation indicators for the quality of fused images.
These are: Spatial Frequency (SF) ~\cite{eskicioglu1995image}, Entropy (EN) ~\cite{roberts2008assessment}, quality of images (Q$_{abf}$) ~\cite{xydeas2000objective}, feature mutual information with wavelet transform(FMI$_w$) ~\cite{haghighat2014fast}, multiscale SSIM (MS-SSIM) ~\cite{ma2015perceptual}, feature mutual information with pixel(FMI$_{pixel}$) ~\cite{haghighat2014fast} Standard Deviation of Image (SD) ~\cite{rao1997fibre}, Visual Information Fidelity (VIF)
~\cite{sheikh2006image}, and mutual information (MI)~\cite{qu2002information}, respectively. In Table.\ref{table:result}, We compared the performance of all methods on 9 evaluation indicators. The best three results are highlighted in {\color{red}{\textbf{red}}}, {\color{brown}{\textbf{brown}}} and {\color{blue}{\textbf{blue}}} fonts.
Our method performed best on 7 indicators and also achieved third place on the remaining two indicators.
Through subjective and objective evaluation, our method is proved to have obvious advantages in performance.

\subsection{Ablation Study}
\textbf{GAN.}
Adversarial learning during training is very effective in image generation tasks, but how to combine it with fusion tasks is a problem in its application.
Our original method only has the generation part of the fused image and does not include two discriminators.
In this case, our method has surpassed the previous method in most objective evaluation indicators.
In order to enhance the characteristics of the fused image: the high resolution of the visible image and the highlighted part of the infrared image, we introduce adversarial learning into the training process.
We use the pre-trained VGG-16 network as a discriminator to enhance the characteristics of different modalities at the feature level.
The objective evaluation results are shown in the Table. \ref{table:gan}. 
Compared with the method that does not use adversarial training, the new method with GAN has improved on seven indicators.
This also proves the effectiveness of introducing generative confrontation methods.

\textbf{Transformer Fusion Module.} 
We propose two transformer fusion methods: spatial transformer and channel transformer. They can work alone or in combination with each other.
In Table. \ref{table:trans}, we separately verify the results of using the two transformer fusion modules alone and in combination.
The effect of passing through the channel transformer first and then passing through the space transformer will be better.
We believe that it is more beneficial for fusion to first pay attention to the channel relationship between corresponding blocks in the process of modelling.

\textbf{Position Embedding.}
In our transformer fusion method, position embedding is removed because the category information provided by position embedding is not needed in the fusion task.
However, whether the direct removal of position embedding has an effect on the training of the transformer has not been verified.
Therefore, we train the TGFuse model with and without position embedding respectively.
Comparing the indicators of the fusion results in Table. \ref{table:pe}, we find that removing position embedding has a positive effect on the results.

\textbf{Transformer Module Layers.}
The transformer model we use is a multi-layer encoder model based on ViT. 
The number of encoder layers also has a great impact on performance.
Unlike classification tasks, fusion tasks are less complex and require fewer layers.
But too few layers may also lead to failure of fusion relationship learning.
Therefore, we set different values for experiments to find the number of layers most suitable for the fusion task.
The comparative results of the experiment are shown in the Table. \ref{table:trans-layer}. 
When the number of layers is three, the test result is a meaningless black image.
It may be that too few layers cause the transformer fusion module can not learn the available fusion relationship.
When the number of layers is five, the test result becomes worse. This may be because the fusion relationship learned by the deep transformer fusion module is redundant.
We select the most suitable number of layers (4 layers) based on the experimental results.

\textbf{CNN Layers.}
Firstly, multi-layer CNN is used to extract features from the input image, which can help the transformer module to converge faster.
The number of layers of CNN (that is, the number of “Res-Block”) affects the granularity and depth of the extracted features.
We set different values to experiment to find the most suitable number of CNN layers. The more layers, the more times the image is downsampled. When the image block is too small, the model cannot learn an effective fusion relationship.
As shown in Table. \ref{table:cnn-layer}, when the depth is 4 layers, the model learns the best fusion relationship. 
When the layer is deeper, the resulting image is meaningless black blocks. This means that if the feature block is too small, the fusion module cannot fuse information effectively.

\textbf{CNN Channels.}
As an important dimension of image features, the number of feature channels is also an important factor influencing algorithm performance.
In the process of feature extraction, we get four image features with the same dimensions but different scales. The difference in the number of channels means that the distribution of channel dimension information is different.
In the ablation experiment, we choose a few typical values as the number of channels. 
After comparison in Table. \ref{table:cnn-channel}, we select the number of channels (64 channels) with the best performance.

\section{Conclusion}
In this paper, we proposed an infrared and visible image fusion method based on a lightweight transformer module and generative adversarial learning. 
The proposed transformer is deeply involved in the fusion task as a fusion relation learning module. 
Adversarial learning provides generators with different modal characteristics during the training process at the feature level.
This is the first attempt of deep combination and application of transformer and adversarial learning in the image fusion task. 
Our method has also achieved outstanding performance in subjective and objective evaluation, which proves the effectiveness and advancement of our method.

\bibliographystyle{IEEEtran}
\bibliography{egbib.bib}

\end{document}